\documentclass[conference]{IEEEtran}
\IEEEoverridecommandlockouts

\newcommand{\DB}[1][]{anonymized-DB } 
\usepackage{array}

\usepackage{cite}
\usepackage{amsmath,amssymb,amsfonts}
\usepackage{algorithmic}
\usepackage{graphicx}
\usepackage{textcomp}
\usepackage{xcolor}
\def\BibTeX{{\rm B\kern-.05em{\sc i\kern-.025em b}\kern-.08em
    T\kern-.1667em\lower.7ex\hbox{E}\kern-.125emX}}
\begin{document}

\title{PAD-Phys: Exploiting Physiology for Presentation Attack Detection in Face Biometrics
}

\author{\IEEEauthorblockN{
Luis~F.~Gomez\IEEEauthorrefmark{1}, 
Julian~Fierrez\IEEEauthorrefmark{1}, 
Aythami~Morales\IEEEauthorrefmark{1},
Mahdi~Ghafourian\IEEEauthorrefmark{1},
Ruben Tolosana\IEEEauthorrefmark{1}, \\
Imanol Solano\IEEEauthorrefmark{1}\IEEEauthorrefmark{2}, 
Alejandro Garcia\IEEEauthorrefmark{2} and
Francisco Zamora-Martinez\IEEEauthorrefmark{2}}

\IEEEauthorblockA{\IEEEauthorrefmark{1}School of Engineering, Universidad Autonoma de Madrid, Spain\\
\{luisf.gomez, julian.fierrez, aythami.morales, mahdi.ghafourian, ruben.tolosana\}@uam.es}

\IEEEauthorblockA{\IEEEauthorrefmark{2}Veridas Digital Authentication Solutions, Pamplona, Spain\\
\{isonalo, agarcia, pzamora\}@veridas.com}

}

\maketitle

\begin{abstract}
Presentation Attack Detection (PAD) is a crucial stage in facial recognition systems to avoid leakage of personal information or spoofing of identity to entities. Recently, pulse detection based on remote photoplethysmography (rPPG) has been shown to be effective in face presentation attack detection. 
This work presents three different approaches to the presentation attack detection based on rPPG: (i) The physiological domain, a domain using rPPG-based models, (ii) the Deepfakes domain, a domain where models were retrained from the physiological domain to specific Deepfakes detection tasks; and (iii) a new Presentation Attack domain was trained by applying transfer learning from the two previous domains to improve the capability to differentiate between bona-fides and attacks. 
The results show the efficiency of the rPPG-based models for presentation attack detection, evidencing a 21.70\% decrease in average classification error rate (ACER) (from 41.03\% to 19.32\%) when the presentation attack domain is compared to the physiological and Deepfakes domains. Our experiments highlight the efficiency of transfer learning in rPPG-based models and perform well in presentation attack detection in instruments that do not allow copying of this physiological feature.

\end{abstract}


\section{Introduction} \label{sec:intro}

Face recognition systems are widely used worldwide in smartphones, e-commerce, or e-bank identification access control systems. This series of face recognition applications has attracted the attention of attackers aiming to exploit system vulnerabilities~\cite{barrero13PRLmultimodalAttack,galbally14reviewAntispoofingFace,hadid15SPMspoofing,2020_JSTSP_GANprintR_Neves,2022_Book-IntroManipulation,2023_Book-PAD_Face_JHO}. 
Vulnerabilities in face recognition systems start with the ability and facility to find and study the users to attack efficiently. Mainly the use of social networks has made it easier to find and select photographs of users with a good definition, which allows the creation of presentation attack instruments such as paper, masks, and replay devices to try to breach such systems~\cite{ghafourian2023toward}.

In order to prevent attackers from having a chance of success, the scientific community has in recent years implemented techniques of presentation attack detection (PAD) to avoid these vulnerabilities in the systems using machine learning~\cite{pan2007eyeblink, liu2018remote} and deep learning~\cite{fang2022learnable, liu2018learning, yu2021transrppg}. Presentation attack detection can be divided into three classes: appearance-based approach, motion-based approach, and remote photoplethysmography-based approach (rPPG-based approach).

Our work is focused on improving biometric presentation attack detection with rPPG-based models and deep learning. 
The main contributions of this work are: 

\begin{itemize}

\item An improved rPPG-based system for face presentation attack detection through domain adaptation across three different approaches: Physiological domain, Deepfakes domain, and a new domain that focuses on presentation attacks created by transfer learning from the first two domains.

\item An experimental study analyzing a novel dataset of videos collected by Veridas, which contains a wide variety of biometric attack instruments and user identities under controlled conditions.

\end{itemize}

The rest of the paper is organized as follows: Related Works provides an overview of the literature on presentation attack detection. 
Materials and Methods present the experimental framework, including the description of the datasets and the methods. 
Experiments and Results summarize the experiments and results. 
Finally, the discussion, conclusions, and future work are drawn in the Discussion and Conclusion.

\subsection{Related Works}

\begin{figure*}[ht]
    \centering
    \includegraphics[width=0.99\textwidth]{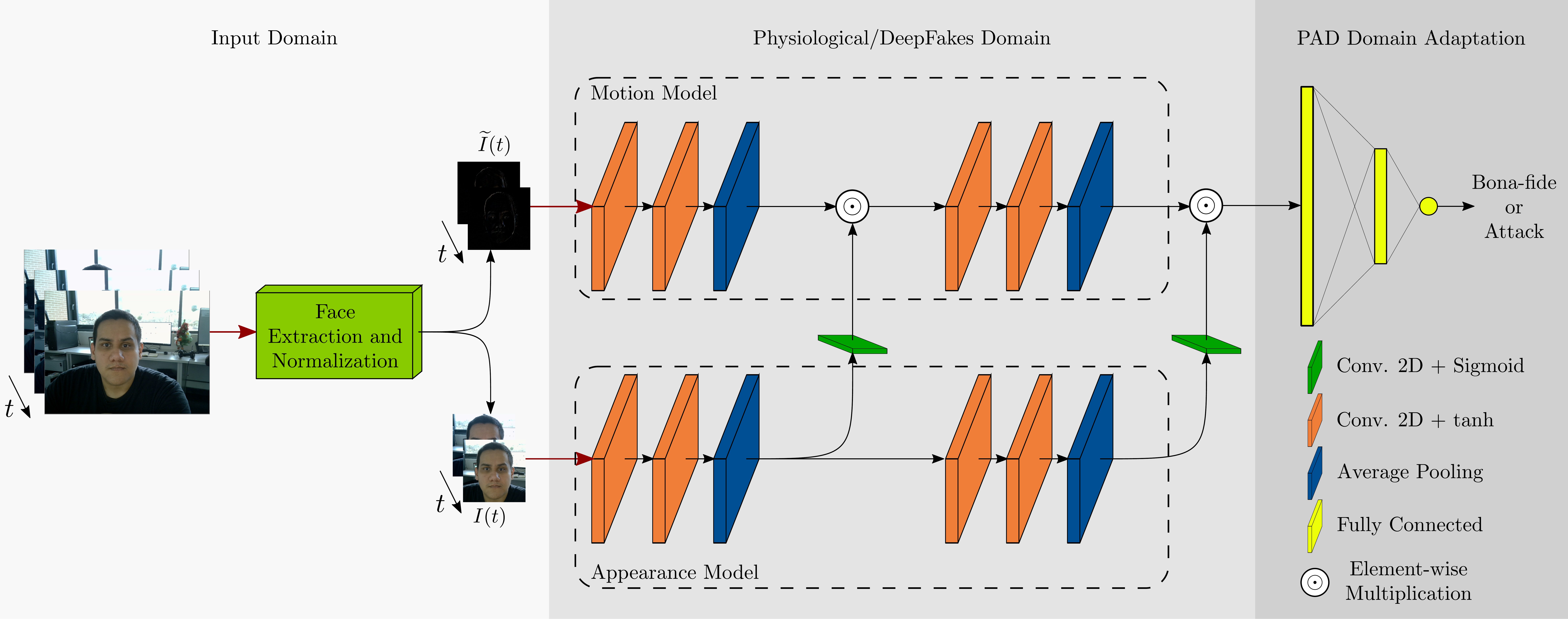}
    \caption{Experimental Framework. It comprises three stages: First, a preprocessing step to normalize the video frames. Then, a Convolutional Attention Network composed of Motion and Appearance Models. Finally, a new classification layer for Presentation Attack Detection.}
    \label{fig:Methods}
\end{figure*}

Face Presentation Attack Detection (PAD) methods that can be divided into three categories:

\subsubsection{Appearance-based methods} 

Appearance-based methods are focused on differentiating bona-fide and attacks from spatial information in the images. The classification can be possible by analyzing color texture~\cite{boulkenafet2016face}, image quality~\cite{galbally2013image,2022_CSUR_FaceQsurvey_Torsten}, and reflection patterns~\cite{wen2015face}; On the other hand, recent studies employ Convolutional neural networks (CNN) in single image analysis~\cite{menotti2015deep, fang2022learnable} or with a combination of additional images provided by invisible lights, infrared cameras, or thermic cameras~\cite{agarwal2017face}. Using additional images generates some difficulties because additional hardware is required to improve the results. 

\subsubsection{Motion-based methods}

Motion-based methods aim to exploit the inconsistencies over time windows on video frames; they try to analyze features such as optical flow~\cite{bao2009liveness}, eye-blink~\cite{pan2007eyeblink}, or facial movements~\cite{pan2007eyeblink, freitas2014face, chetty2006multi}. These methods may not work in specific attacks that expose the attacker's eyes and mouth~\cite{zhang2011face} or when the attacker uses video replay strategies.

\subsubsection{rPPG-based methods} \label{sec:rPPG}

rPPG-based methods are a non-contact technique for recovering physiological signals through a conventional RGB camera under an environmental light through the analysis of changes in the skin color pigmentation caused by heartbeats~\cite{mcduff2015survey,  chen2018deepphys, hernandez2020comparative, estepp2014recovering}.

Traditional rPPG-based methods employ the extraction of temporal signals per channel using all the image, the face, or specific regions of interest over the face like the cheeks or the forehead. After that, the spectrum, the energy, or the total power is used as features in PAD~\cite{li2016generalized, liu2018remote, liu20163d, niu2020video, 2018_CVPRW_timePulsePAD_JH}. 
On the other hand, Deep rPPG-based methods employ a combination of appearance and motion methods, where the appearance features can be extracted with pretrained CNN and later use a Recurrent Neural Network (RNN) to find a motion/temporal correlation between the features. Another approach is using Visual Transformers (ViT) to make the appearance and motion correlation in one module~\cite{liu2018learning, liu2022learning, yu2021transrppg}.

rPPG-based methods are a solution that works on conventional presentation attacks such as paper attacks and 3D mask attacks because these materials do not show heartbeat signals. Otherwise, when the attacks are carried out with video replays, these attacks can capture the heartbeat patterns of the user if the video is recorded with enough quality.

\section{Materials and methods} \label{sec:Methods}

We propose an experimental framework where physiological information is explored at different levels (see Figure~\ref{fig:Methods}). The hypotheses and experiment proposed are presented below.

\vspace*{3mm}

    \noindent \textbf{Physiological Domain (Level 1).} We propose to use a rPPG prediction models to obtain physiological information.

    \begin{itemize}
    \item \emph{Hypothesis (H1):} videos captured provide physiological information that allows us presentation attack detection.
    \item \emph{Experiment:} we evaluate presentation attack detection performance for different bona-fide and biometric presentation attack videos using the DeepPhys model~\cite{hernandez2020deepfakeson}. 
    \end{itemize}

    \noindent \textbf{DeepFakes Domain (Level 2).} We propose to evaluate different DeepFakes detection models \cite{2020_INFFUS_SurveyDeepFakes_Tolosana} adapted from physiological domains.

    \begin{itemize}
    \item \emph{Hypothesis (H2):} models based on DeepFakes detection provide additional information to improve the PAD based on rPPG models.
    \item \emph{Experiment:} the rPPG models (\textit{H1}) are used as pre-trained models and then finetuned with two DeepFakes databases (Celeb-DF v2 and DFDC preview databases). The resulting models are used to evaluate bona-fide and biometric presentation attack videos.
    \end{itemize}

    \noindent \textbf{Presentation Attacks Domain (Level 3).} We propose to improve presentation attack detection by incorporating a private database with multiple presentation attack instruments.

    \begin{itemize}
    \item \emph{Hypothesis (H3):} Presentation attack detection is improved when presentation attack instruments are incorporated into the training process.
    
    \item \emph{Experiment:} rPPG (\textit{H1}) and DeepFakes (\textit{H2}) models are adapted to presentation attacks domain using a private database with different presentation attack instruments.
\end{itemize}
Details of the methods implemented to validate all hypotheses are presented in Methods.

\subsection{Databases} \label{Databases}

\subsubsection{Celeb-DF v2 Database} 

Celeb-DF v2 Database was created to get a better visual quality compared with other databases of DeepFakes~\cite{li2020celeb,2022_EAAI_DeepFakes_Tolosana}. Fifty-nine subjects are chosen with various distribution in their genders, ages, and ethnic groups. The database includes 590 real videos and 5639 fake videos, where the real videos were recollected from YouTube, and the fake videos were created using swapping faces between pairs of subjects. 

\subsubsection{DFDC Preview} 

Facebook released the DFDC database~\cite{dolhansky2019deepfake,2022_EAAI_DeepFakes_Tolosana} in collaboration with other companies and academic institutions. In this work, we consider the DFDC Preview dataset, composed of $5$.$214$ videos ($1$.$131$ real videos and $4$.$083$ fake videos) from 66 subjects. The videos have diversity in gender, skin tone, age, illuminations, and backgrounds. Fake videos were generated by swapping faces with similar attributes such as
skin-tone, facial hair, and glasses.

\subsubsection{Veridas DB} 

Veridas created the Veridas database. The database contains $10$.$397$  video recordings of 53 users, and it is divided into 11 attack types. Table~\ref{tab:DB_Description} summarizes the number of videos per attack in Veridas DB.
Veridas DB was split using a double-stratified split between the presentation attack instruments and user identities. The $10.397$ videos from Veridas DB were divided into $2.649$, $662$, and $7.086$ for training, validation, and testing, respectively. All the videos in Veridas DB were recorded while the users performed head movements, including looking to the left, right, up, and down in front of the camera. These movements enable the application of motion-based PAD methods. 

\begin{table}[t]
\centering
\caption{Number of videos for each attack and number of bona-fides videos in Veridas DB.}
\label{tab:DB_Description}
\resizebox{0.9\columnwidth}{!}{%
\begin{tabular}{lcc}
\noalign{\hrule height 1pt}
Veridas DB        & Abbreviation & Videos \\
\noalign{\hrule height 1pt}
Bona-fide                    & BF    & 2061   \\
Mannequin Head Attack        & MH    & 1332   \\
Layered 2D Transparent Photo & L2TP  & 1308   \\
Plastic Mask Attack          & PlM   & 1303   \\
Latex Mask Attack            & LM    & 1190   \\
3D Curved Paper Mask         & 3CPM  & 1162   \\
Print Mask Attack            & PrM   & 711    \\
Video Replay Attack          & VR    & 489    \\
Print Attack                 & Pr    & 244    \\
Silicone Mask Attack         & SM    & 240    \\
Photo Replay Attack          & PR    & 189    \\
Print 3D Layered Mask        & Pr3LM        & 144    \\
Total attacks                & --   & 8336   \\
\noalign{\hrule height 1pt}
\end{tabular}%
}
\end{table}

\begin{table}[t]
\caption{Comparison table between the different databases used in this work.}
\label{tab:DBs_Comparation}
\resizebox{\columnwidth}{!}{%
\begin{tabular}{lcccc}
\noalign{\hrule height 1pt}
         & \#Videos & Bona-fides & Attacks & Users \\
 \noalign{\hrule height 1pt}
Celeb-DF~\cite{li2020celeb} & $5214$     & $1131$      & $4083$   & $66$    \\
DFDC~\cite{dolhansky2019deepfake}      & $6229$     & $590$       & $5639$   & $59$    \\
Veridas DB      & $10397$    & $2061$      & $8336$   & $53$   \\
\noalign{\hrule height 1pt}
\end{tabular}%
}
\end{table}

Table~\ref{tab:DBs_Comparation} finally shows a summary of the key numbers in the 3 above-described datasets used in the present paper: number of videos, division between bona-fides and attacks, and number of users.

\newpage

\subsection{Methods} \label{Methods}

\subsubsection{Face extraction}

The videos were preprocessed using a face detection engine based on MobileNet v2, fine-tuned for face detection. The square bounding box obtained is expanded by 80\% about the bounding box width. Additionally, one exponential moving average (EMA) was used to smooth the changes in the bounding box over the video. Finally, each cropped face $I(t)$ was resized to 36$\times$36$\times$3 pixels.

\subsubsection{DeepPhys} \label{sec:DeepPhys}

In this work, we employ the DeepPhys model, a Convolutional Attention Network created by Chen and McDuff in~\cite{chen2018deepphys} and implemented by Hernandez-Ortega et al. in~\cite{hernandez2020comparative} where DeepPhys was trained to predict the hearth rate on COHFACE database~\cite{heusch2017reproducible}. DeepPhys's objective is to estimate the human heart rate using photoplethysmography and facial video sequences. The model comprises two parallel Convolutional Neuronal Networks (CNN) that extract spatial and temporal information from videos, combining motion and appearance across the model.
Figure~\ref{fig:Methods} shows the two CNN branches from the DeepPhys model, and they are described below:

\begin{itemize}
    \item \textbf{Motion model:} It is designed to realize a short-time analysis of the videos to detect pixel changes over the scene. Input model at time $t$ is calculated to follow the next equation:
    \begin{equation}
        \widetilde{I}(t) = \frac{I(t) - I(t-1)}{I(t) + I(t-1)}
    \end{equation}
    \item \textbf{Appearance model:} It is designed to create attention masks based on the subject's appearance to help the motion model. Input model at time $t$ is the raw frame $I(t)$ 
\end{itemize}

\subsubsection{DeepFakesON-Phys}  \label{sec:DeepFakes}

DeepFakesON-Phys model was based on DeepPhys and adapted to DeepFakes detection~\cite{hernandez2020deepfakeson}. DeepFakesON-Phys was initialized with the weights of DeepPhys as a starting point and retrained completely with DeepFakes databases. DeepFakesON-Phys has two versions, the first model was trained with the DFDC preview database, and the second model was trained with Celeb-DF v2 Database.

\subsubsection{PAD-Phys} \label{sec:DeepPAD}

PAD-Phys is the first approach to PAD using DeepPhys models. PAD-Phys training was adapted from DeepPhys and DeepFakesON-Phys models, leaving all CNN layers frozen before the flatten layer and retraining two new last layers to classify between bona-fide and attacks. 

PAD-Phys models were trained with the Veridas DB, which includes 11 presentation attack instruments, resulting in three models: (i) PAD-Phys Vr.~1, based on DeepPhys; (ii) PAD-Phys Vr.~2, based on DeepFakesON\_Phys trained with Celeb-DF; and (iii) PAD-Phys Vr.~3, based on DeepFakesON\_Phys trained with DFDC.

\newpage

\subsubsection{Performance metrics}

The final decision score for the video is the mean value over the predicted temporal score obtained after each model \cite{2018_CVPRW_timePulsePAD_JH,2019_BookFFER_ContFacePAD_JHO,2022_Handbook_DeepFakesPhys_JHO}.
Classification results are determined by the threshold obtained from analyzing the equal error rate in the validation data partition.
The performance is reported using metrics such as the attack presentation classification error rate (APCER), the bona-fide presentation classification error rate (BPCER), the average classification error rate (ACER), and the receiver operating characteristic curve (ROC curve).


BPCER is the proportion of bona-fide samples that are misclassified as attacks, equivalent to the False Negative Rate (FNR). APCER, on the other hand, is the proportion of attack samples incorrectly classified as genuine, equivalent to the False Positive Rate (FPR).

\section{Experiments and results}

\begin{figure*}[ht]
    \includegraphics[width=0.99\textwidth]{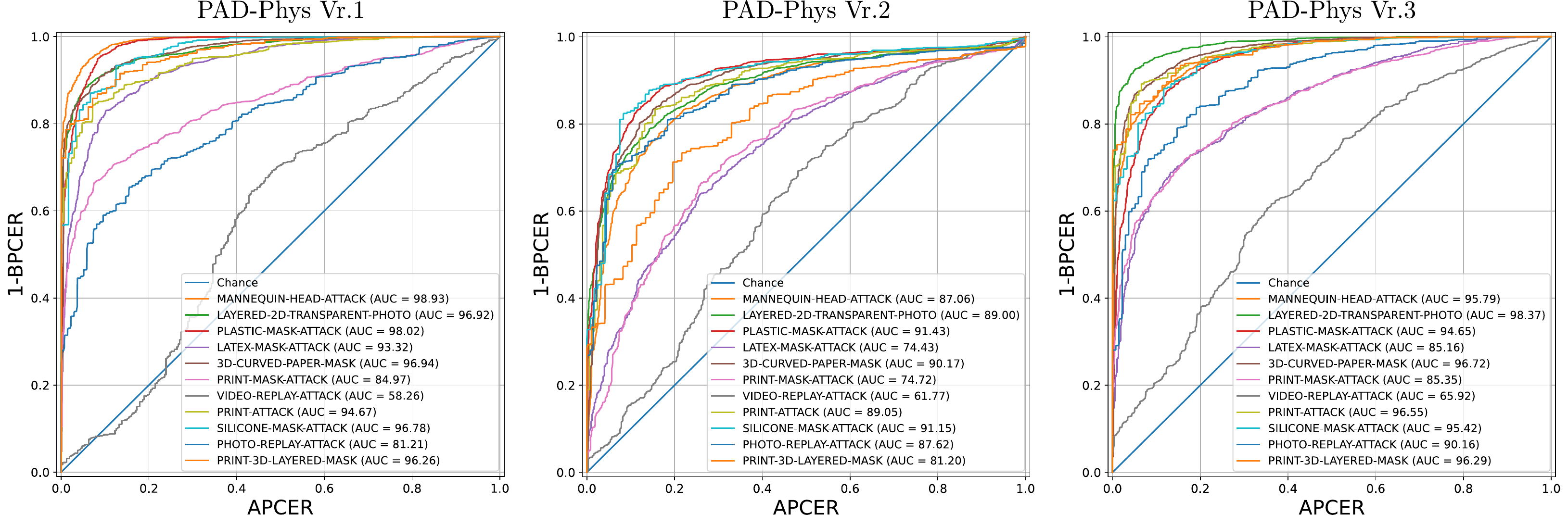}
    \caption{ROC curves for each attack type in Veridas DB using the three models.}
    \label{fig:ROC_BestModels}
\end{figure*}

\subsection{DeepPhys}

Following the methodology presented in Section~\ref{sec:DeepPAD}, the DeepPhys model was retrained with $2.649$, $662$, and $7.086$ videos for training, validation, and testing, respectively, from Veridas DB. 
Table~\ref{tab:Results_Vr1} shows the PAD performance for the DeepPhys model and PAD-Phys Vr.1 (based on DeepPhys) models for the test partition on Veridas DB using a fixed threshold obtained from the validation partition. 
Results show a BPCER of 41.44\% and a BPCER of 18.28\% for DeepPhys and PAD-Phys Vr. 1, respectively.

\begin{table}[t]
\caption{Performance metrics for DeepPhys and PAD-Phys models retrained on each Veridas DB attack for an BPCER of 41.44\% and an BPCER of 18.28\%, respectively in each model.}
\label{tab:Results_Vr1}
\resizebox{\columnwidth}{!}{%
\begin{tabular}{>{\centering\arraybackslash}lwc{0.2\columnwidth}wc{0.2\columnwidth}wc{0.2\columnwidth}wc{0.2\columnwidth}}
\noalign{\hrule height 1pt}
      & \multicolumn{2}{c}{DeepPhys} & \multicolumn{2}{c}{ \makebox[0pt]{PAD-Phys Vr.1}} \\
      & APCER  & ACER  & APCER        & ACER  \\
\noalign{\hrule height 1pt}
MH    & $42.01$  & $41.73$        & $1.06$  & $9.67$  \\
L2TP  & $34.36$  & $37.90$        & $2.57$  & $10.42$        \\
PlM   & $47.36$  & $44.40$        & $3.50$  & $10.89$        \\
LM    & $50.97$  & $46.20$        & $9.69$  & $13.99$        \\
3CPM  & $48.50$  & $44.97$        & $2.28$  & $10.28$        \\
PrM   & $62.32$  & $51.88$        & $32.00$ & $25.14$        \\
VR    & $49.84$  & $45.64$        & $69.50$ & $43.89$        \\
Pr    & $34.91$  & $38.18$        & $7.19$  & $12.73$        \\
SM    & $46.73$  & $44.09$        & $4.17$  & $11.22$        \\
PR    & $32.35$  & $36.90$        & $41.18$ & $29.73$        \\
Pr3LM & $41.67$  & $41.55$        & $6.19$  & $12.23$       \\
\noalign{\hrule height 0.4pt}
Total & $45.33$  & $43.38$        & $10.94$ & $14.61$       \\
\noalign{\hrule height 1pt}
\end{tabular}%
}
\end{table}

Results show that DeepPhys initially has a performance over the chance for all the attacks, with ACER around 43.38\%. 
These results support the first hypothesis (\textit{H1}), which presents the idea that the videos captured provide physiological information to differentiate between bona-fides and attacks.
Additionally, Table~\ref{tab:Results_Vr1} shows a reduction in the ACER for almost all the attacks, obtaining an average ACER decrease of 28.77 percentage points (from 43.38\% to 14.61\%). These results suggest that the third hypothesis (\textit{H3}) is partially accepted.

\newpage

\subsection{DeepFakesON-Phys}

Table~\ref{tab:Results_Vr2} shows the test partition results for the fixed validation threshold where DeepFakesON-Phys Vr. Celeb-DF and PAD-Phys Vr. 2 models obtain BPCER values of 39.60\% and 24.97\%, respectively. Similarly, Table~\ref{tab:Results_Vr3} shows the results for the fixed threshold obtained in the validation partition where DeepFakesON-Phys Vr. DFDC and PAD-Phys Vr. 3 models obtain BPCER values of 42.45\% and 29.03\%, respectively.

\begin{table}[t]
\caption{Performance metrics for the DeepFakes pretrained on Celeb-DF and PAD-Phys models retrained on each Veridas DB attack for an BPCER of 39.60\% and an BPCER of 24.97\%, respectively.}
\label{tab:Results_Vr2}
\resizebox{\columnwidth}{!}{%
\begin{tabular}{>{\centering\arraybackslash}lwc{0.2\columnwidth}wc{0.2\columnwidth}wc{0.2\columnwidth}wc{0.2\columnwidth}}
\noalign{\hrule height 1pt}
      & \multicolumn{2}{c}{DeepFakesON\_Phys Vr.   Celeb-DF} & \multicolumn{2}{c}{PAD-Phys Vr.2} \\
      & APCER  & ACER  & APCER        & ACER  \\
\noalign{\hrule height 1pt}
MH    & $41.67$       & $40.63$      & $14.24$        & $19.60 $       \\
L2TP  & $29.24$       & $34.42$      & $10.85$        & $17.91$        \\
PlM   & $29.34$       & $34.47$      & $7.32$         & $16.14 $       \\
LM    & $34.58$       & $37.09$      & $39.71$        & $32.34$        \\
3CPM  & $41.76$       & $40.68 $     & $9.70$         & $17.33$        \\
PrM   & $56.63$       & $48.11$      & $36.63$        & $30.80$        \\
VR    & $48.89$       & $44.24$      & $55.66$        & $40.31$        \\
Pr    & $50.30$       & $44.95$      & $11.98$        & $18.47$        \\
SM    & $49.93$       & $44.76$      & $7.50$         & $16.23$        \\
PR    & $57.35$       & $48.47$      & $13.97$        & $19.47$        \\
Pr3LM & $50.00$       & $44.80$      & $29.90$        & $27.43$       \\
\noalign{\hrule height 0.4pt}
Total & $38.79$       & $39.19$      & $22.01$        & $23.49$  \\
\noalign{\hrule height 1pt}
\end{tabular}%
}
\end{table}

\begin{table}[t]
\caption{Performance metrics for the DeepFakes pretrained on DFDC and PAD-Phys models retrained on each Veridas DB attack for an BPCER of 42.45\% and an BPCER of 29.03\%, respectively.}
\label{tab:Results_Vr3}
\resizebox{\columnwidth}{!}{%
\begin{tabular}{>{\centering\arraybackslash}lwc{0.2\columnwidth}wc{0.2\columnwidth}wc{0.2\columnwidth}wc{0.2\columnwidth}}
\noalign{\hrule height 1pt}
      & \multicolumn{2}{c}{DeepFakesON\_Phys Vr.   DFDC} & \multicolumn{2}{c}{PAD-Phys Vr.3} \\
      & APCER  & ACER  & APCER        & ACER         \\
\noalign{\hrule height 1pt}
MH    & $45.49$     & $43.97$    & $2.35$         & $15.69$  \\
L2TP  & $13.54$     & $27.99$    & $0.34$         & $14.69$  \\
PlM   & $48.54$     & $45.50$    & $4.81$         & $16.92$  \\
LM    & $51.10$     & $46.77$    & $15.94$        & $22.49$  \\
3CPM  & $31.29$     & $36.87$    & $1.56$         & $15.30$  \\
PrM   & $37.05$     & $39.75$    & $16.00$        & $22.52$  \\
VR    & $38.41$     & $40.43$    & $47.48$        & $38.26$  \\
Pr    & $29.59$     & $36.02$    & $1.80$         & $15.42$  \\
SM    & $50.51$     & $46.48$    & $3.33$         & $16.18$  \\
PR    & $12.50$     & $27.47$    & $8.09$         & $18.56$  \\
Pr3LM & $46.88$     & $44.66$    & $0.00$         & $14.52$  \\
\noalign{\hrule height 0.4pt}
Total & $38.60$     & $40.52$    & $10.71$        & $19.87$  \\
\noalign{\hrule height 1pt}
\end{tabular}%
}
\end{table}


DeepFakesON-Phys models show a slight improvement in ACER values of around 39.85\%, in L2TP, PR, PrM, and 3CPM attacks decreasing by around 9.11\% in ACER.
This performance shows that DeepFakes detection models give us additional information for specific PAD and supports the second hypothesis (\textit{H2}).


Also, the results obtained from the PAD-Phys models shows a reduction in the ACER for almost all the attacks, obtaining an average ACER decrease of 15.7\% and 20.65\% for PAD-Phys Vr.~2 and Vr.~3, respectively.
Based on these results, the third hypothesis (\textit{H3}) proposed can be accepted.

\newpage

\section{Conclusions / Discussions}

This work explores the performance of models based on remote photoplethysmography for presentation attacks detection. 
Videos with 11 types of attacks and diverse identities were considered for this study. 
Our first approach, using trained models for pulse detection and DeepFakes detection, showed results not far from randomness with 43.38\%, 39.19\%, and 40.52\% in average ACER (See Table~\ref{tab:Results_Means}), evidencing that physiological feature-based models could be helpful in this problem. 
These results allow us to propose new models based on rPPG and DeepFakes to be retrained for biometric presentation attacks.

\begin{table}[t]
\caption{Average BPCER, APCER and ACER obtained by all the models used in this work.}
\label{tab:Results_Means}
\resizebox{\columnwidth}{!}{%
\begin{tabular}{lccc}
\noalign{\hrule height 1pt}
         & BPCER & APCER & ACER  \\
\noalign{\hrule height 1pt}
DeepPhys~\cite{chen2018deepphys}        & $41.44$ & $45.33$ & $43.38$ \\
DeepFakesON\_Phys Vr. Celeb-DF~\cite{hernandez2020deepfakeson}   & $39.60$ & $38.79$ & $39.19$ \\
DeepFakesON\_Phys Vr. DFDC~\cite{hernandez2020deepfakeson}     & $42.45$ & $38.60$ & $40.52$ \\
PAD-Phys Vr.1   & $18.28$ & $10.94$ & $14.61$ \\
PAD-Phys Vr.2          & $24.97$ & $22.01$ & $23.49$ \\
PAD-Phys Vr.3          & $29.03$ & $10.71$ & $19.87$ \\
\noalign{\hrule height 1pt}
\end{tabular}%
}
\end{table}

PAD-Phys models get better results for each presentation attack instrument, with one exception in the VR attack, where the performance in the three models does not significantly change (see Figure~\ref{fig:ROC_BestModels}). This attack is the only biometric presentation attack in which the replayed user's video can still have physiological information that confuses the model in the presentation attack detection, a problem already exposed in the Related Work (see Section~\ref{sec:rPPG}).
Our models show an overall increase in most attacks by an average ACER of 21.70\% compared with the original models. 


These results suggest that most of the biometric presentation attacks in the database can be detected thanks to the fact that the used attacks do not contain real pulse presence as in presentation attack instruments such as paper, masks, mannequins, and screens; except for the video-replay attack where the PAD may be confused due to the real-time playback of the user, and this may contain physiological information.

\begin{figure}[t!]
    \centering
    \includegraphics[width=0.95\columnwidth]{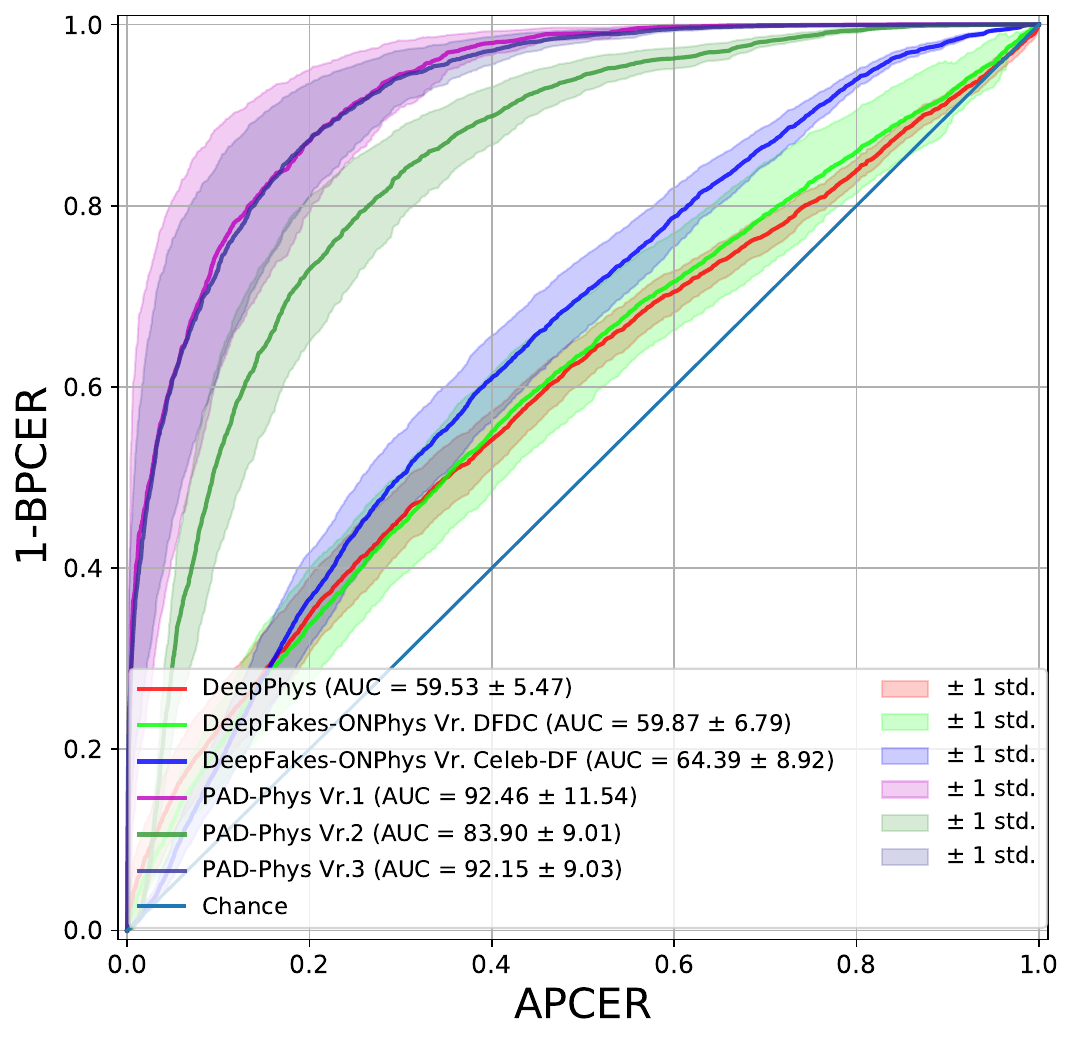}
    \caption{ROC curve - Veridas DB attacks. Comparison between the different domain adaptation strategies.}
    \label{fig:ROC_means}
\end{figure} 


Finally, Figure~\ref{fig:ROC_means} shows the mean ROC for all the models used in this work, evidencing the improvement in the performance of the models in presentation attack detection; this encourages us to continue exploring new approaches based on physiological characteristics exploiting new learning architectures such as transformers.

\section{Acknowledgments}
This work has been supported by Veridas and by projects: TRESPASS-ETN (ITN-2019-860813), PRIMA (ITN2019-860315), and
BBforTAI (PID2021-127641OB-I00 MICINN/FEDER).

\newpage

\bibliographystyle{IEEEtran}
\bibliography{IEEEabrv,mybibfile}

\begin{thebibliography}{10}
\providecommand{\url}[1]{#1}
\csname url@samestyle\endcsname
\providecommand{\newblock}{\relax}
\providecommand{\bibinfo}[2]{#2}
\providecommand{\BIBentrySTDinterwordspacing}{\spaceskip=0pt\relax}
\providecommand{\BIBentryALTinterwordstretchfactor}{4}
\providecommand{\BIBentryALTinterwordspacing}{\spaceskip=\fontdimen2\font plus
\BIBentryALTinterwordstretchfactor\fontdimen3\font minus
  \fontdimen4\font\relax}
\providecommand{\BIBforeignlanguage}[2]{{%
\expandafter\ifx\csname l@#1\endcsname\relax
\typeout{** WARNING: IEEEtran.bst: No hyphenation pattern has been}%
\typeout{** loaded for the language `#1'. Using the pattern for}%
\typeout{** the default language instead.}%
\else
\language=\csname l@#1\endcsname
\fi
#2}}
\providecommand{\BIBdecl}{\relax}
\BIBdecl

\bibitem{barrero13PRLmultimodalAttack}
M.~Gomez-Barrero, J.~Galbally, and J.~Fierrez, ``Efficient software attack to
  multimodal biometric systems and its application to face and iris fusion,''
  \emph{Pattern Recognition Letters}, vol.~36, pp. 243--253, January 2014.

\bibitem{galbally14reviewAntispoofingFace}
J.~Galbally, S.~Marcel, and J.~Fierrez, ``Biometric anti-spoofing methods: A
  survey in face recognition,'' \emph{IEEE Access}, vol.~2, pp. 1530--1552,
  December 2014.

\bibitem{hadid15SPMspoofing}
A.~Hadid, N.~Evans, S.~Marcel, and J.~Fierrez, ``Biometrics systems under
  spoofing attack: an evaluation methodology and lessons learned,'' \emph{IEEE
  Signal Processing Magazine}, vol.~32, no.~5, pp. 20--30, September 2015.

\bibitem{2020_JSTSP_GANprintR_Neves}
J.~C. Neves, R.~Tolosana, R.~Vera-Rodriguez, V.~Lopes, H.~Proenca, and
  J.~Fierrez, ``{GANprintR}: Improved fakes and evaluation of the state of the
  art in face manipulation detection,'' \emph{IEEE Journal of Selected Topics
  in Signal Processing}, vol.~14, no.~5, pp. 1038--1048, August 2020.

\bibitem{2022_Book-IntroManipulation}
R.~Tolosana, R.~Vera-Rodriguez, J.~Fierrez, A.~Morales, and J.~Ortega-Garcia,
  \emph{Handbook of Digital Face Manipulation and Detection}.\hskip 1em plus
  0.5em minus 0.4em\relax Springer, 2022, ch. An Introduction to Digital Face
  Manipulation, pp. 3--26.

\bibitem{2023_Book-PAD_Face_JHO}
J.~Hernandez-Ortega, J.~Fierrez, A.~Morales, and J.~Galbally, \emph{Handbook of
  Biometric Anti-Spoofing}.\hskip 1em plus 0.5em minus 0.4em\relax Springer,
  2023, ch. Introduction to Presentation Attack Detection in Face Biometrics
  and Recent Advances, p. 203–230, 3rd Ed.

\bibitem{ghafourian2023toward}
M.~Ghafourian, J.~Fierrez, L.~F. Gomez, R.~Vera-Rodriguez, A.~Morales,
  Z.~Rezgui, and R.~Veldhuis, ``Toward face biometric de-identification using
  adversarial examples,'' in \emph{AAAI-23 Workshop on Artificial Intelligence
  for Cyber Security (AICS)}, 2023.

\bibitem{pan2007eyeblink}
G.~Pan, L.~Sun, Z.~Wu, and S.~Lao, ``Eyeblink-based anti-spoofing in face
  recognition from a generic webcamera,'' in \emph{2007 IEEE 11th International
  Conference on Computer Vision}.\hskip 1em plus 0.5em minus 0.4em\relax IEEE,
  2007, pp. 1--8.

\bibitem{liu2018remote}
S.-Q. Liu, X.~Lan, and P.~C. Yuen, ``Remote photoplethysmography correspondence
  feature for {3D} mask face presentation attack detection,'' in
  \emph{Proceedings of the European Conference on Computer Vision (ECCV)},
  2018, pp. 558--573.

\bibitem{fang2022learnable}
M.~Fang, N.~Damer, F.~Kirchbuchner, and A.~Kuijper, ``Learnable multi-level
  frequency decomposition and hierarchical attention mechanism for generalized
  face presentation attack detection,'' in \emph{Proceedings of the IEEE/CVF
  Winter Conference on Applications of Computer Vision (WACV)}, 2022, pp.
  3722--3731.

\bibitem{liu2018learning}
Y.~Liu, A.~Jourabloo, and X.~Liu, ``Learning deep models for face
  anti-spoofing: Binary or auxiliary supervision,'' in \emph{Proceedings of the
  IEEE Conference on Computer Vision and Pattern Recognition (CVPR)}, 2018, pp.
  389--398.

\bibitem{yu2021transrppg}
Z.~Yu, X.~Li, P.~Wang, and G.~Zhao, ``{TransRPPG}: Remote photoplethysmography
  transformer for {3D} mask face presentation attack detection,'' \emph{IEEE
  Signal Processing Letters}, vol.~28, pp. 1290--1294, 2021.

\bibitem{boulkenafet2016face}
Z.~Boulkenafet, J.~Komulainen, and A.~Hadid, ``Face spoofing detection using
  colour texture analysis,'' \emph{IEEE Transactions on Information Forensics
  and Security}, vol.~11, no.~8, pp. 1818--1830, 2016.

\bibitem{galbally2013image}
J.~Galbally, S.~Marcel, and J.~Fierrez, ``Image quality assessment for fake
  biometric detection: Application to iris, fingerprint, and face
  recognition,'' \emph{IEEE Transactions on Image Processing}, vol.~23, no.~2,
  pp. 710--724, 2013.

\bibitem{2022_CSUR_FaceQsurvey_Torsten}
T.~Schlett, C.~Rathgeb, O.~Henniger, J.~Galbally, J.~Fierrez, and C.~Busch,
  ``Face image quality assessment: A literature survey,'' \emph{ACM Computing
  Surveys}, vol.~54, no.~10, pp. 1--49, 2022.

\bibitem{wen2015face}
D.~Wen, H.~Han, and A.~K. Jain, ``Face spoof detection with image distortion
  analysis,'' \emph{IEEE Transactions on Information Forensics and Security},
  vol.~10, no.~4, pp. 746--761, 2015.

\bibitem{menotti2015deep}
D.~Menotti, G.~Chiachia, A.~Pinto, W.~R. Schwartz, H.~Pedrini, A.~X. Falcao,
  and A.~Rocha, ``Deep representations for iris, face, and fingerprint spoofing
  detection,'' \emph{IEEE Transactions on Information Forensics and Security},
  vol.~10, no.~4, pp. 864--879, 2015.

\bibitem{agarwal2017face}
A.~Agarwal, D.~Yadav, N.~Kohli, R.~Singh, M.~Vatsa, and A.~Noore, ``Face
  presentation attack with latex masks in multispectral videos,'' in
  \emph{Proceedings of the IEEE Conference on Computer Vision and Pattern
  Recognition Workshops (CVPRw)}, 2017, pp. 81--89.

\bibitem{bao2009liveness}
W.~Bao, H.~Li, N.~Li, and W.~Jiang, ``A liveness detection method for face
  recognition based on optical flow field,'' in \emph{2009 International
  Conference on Image Analysis and Signal Processing}.\hskip 1em plus 0.5em
  minus 0.4em\relax IEEE, 2009, pp. 233--236.

\bibitem{freitas2014face}
T.~d. Freitas~Pereira, J.~Komulainen, A.~Anjos, J.~M. De~Martino, A.~Hadid,
  M.~Pietik{\"a}inen, and S.~Marcel, ``Face liveness detection using dynamic
  texture,'' \emph{EURASIP Journal on Image and Video Processing}, vol. 2014,
  pp. 1--15, 2014.

\bibitem{chetty2006multi}
G.~Chetty and M.~Wagner, ``Multi-level liveness verification for face-voice
  biometric authentication,'' in \emph{2006 Biometrics Symposium: Special
  Session on Research at the Biometric Consortium Conference}.\hskip 1em plus
  0.5em minus 0.4em\relax IEEE, 2006, pp. 1--6.

\bibitem{zhang2011face}
Z.~Zhang, D.~Yi, Z.~Lei, and S.~Z. Li, ``Face liveness detection by learning
  multispectral reflectance distributions,'' in \emph{2011 IEEE International
  Conference on Automatic Face \& Gesture Recognition (FG)}.\hskip 1em plus
  0.5em minus 0.4em\relax IEEE, 2011, pp. 436--441.

\bibitem{mcduff2015survey}
D.~J. McDuff, J.~R. Estepp, A.~M. Piasecki, and E.~B. Blackford, ``A survey of
  remote optical photoplethysmographic imaging methods,'' in \emph{2015 37th
  Annual International Conference of the IEEE Engineering in Medicine and
  Biology Society (EMBC)}.\hskip 1em plus 0.5em minus 0.4em\relax IEEE, 2015,
  pp. 6398--6404.

\bibitem{chen2018deepphys}
W.~Chen and D.~McDuff, ``{DeepPhys}: Video-based physiological measurement
  using convolutional attention networks,'' in \emph{Proceedings of the
  European Conference on Computer Vision (ECCV)}, 2018, pp. 349--365.

\bibitem{hernandez2020comparative}
J.~Hernandez-Ortega, J.~Fierrez, A.~Morales, and D.~Diaz, ``A comparative
  evaluation of heart rate estimation methods using face videos,'' in
  \emph{IEEE Conf. on Computers, Software, and Applications (COMPSAC)}, July
  2020, pp. 1438--1443.

\bibitem{estepp2014recovering}
J.~R. Estepp, E.~B. Blackford, and C.~M. Meier, ``Recovering pulse rate during
  motion artifact with a multi-imager array for non-contact imaging
  photoplethysmography,'' in \emph{2014 IEEE International Conference on
  Systems, Man, and Cybernetics (SMC)}.\hskip 1em plus 0.5em minus 0.4em\relax
  IEEE, 2014, pp. 1462--1469.

\bibitem{li2016generalized}
X.~Li, J.~Komulainen, G.~Zhao, P.-C. Yuen, and M.~Pietik{\"a}inen,
  ``Generalized face anti-spoofing by detecting pulse from face videos,'' in
  \emph{2016 23rd International Conference on Pattern Recognition
  (ICPR)}.\hskip 1em plus 0.5em minus 0.4em\relax IEEE, 2016, pp. 4244--4249.

\bibitem{liu20163d}
S.~Liu, P.~C. Yuen, S.~Zhang, and G.~Zhao, ``{3D} mask face anti-spoofing with
  remote photoplethysmography,'' in \emph{Proceedings of the European
  Conference on Computer Vision (ECCV)}.\hskip 1em plus 0.5em minus 0.4em\relax
  Springer, 2016, pp. 85--100.

\bibitem{niu2020video}
X.~Niu, Z.~Yu, H.~Han, X.~Li, S.~Shan, and G.~Zhao, ``Video-based remote
  physiological measurement via cross-verified feature disentangling,'' in
  \emph{Proceedings of the European Conference on Computer Vision
  (ECCV)}.\hskip 1em plus 0.5em minus 0.4em\relax Springer, 2020, pp. 295--310.

\bibitem{2018_CVPRW_timePulsePAD_JH}
J.~Hernandez-Ortega, J.~Fierrez, A.~Morales, and P.~Tome, ``Time analysis of
  pulse-based face anti-spoofing in visible and nir,'' in \emph{Proc. IEEE
  Conf. on Computer Vision and Pattern Recognition Workshops, CVPRW}, June
  2018.

\bibitem{liu2022learning}
S.-Q. Liu, X.~Lan, and P.~C. Yuen, ``Learning temporal similarity of remote
  photoplethysmography for fast {3D} mask face presentation attack detection,''
  \emph{IEEE Transactions on Information Forensics and Security}, vol.~17, pp.
  3195--3210, 2022.

\bibitem{hernandez2020deepfakeson}
J.~Hernandez-Ortega, R.~Tolosana, J.~Fierrez, and A.~Morales,
  ``{DeepFakesON-Phys}: Deepfakes detection based on heart rate estimation,''
  in \emph{AAAI Workshop on Artificial Intelligence Safety (SafeAI)}, February
  2020.

\bibitem{2020_INFFUS_SurveyDeepFakes_Tolosana}
R.~Tolosana, R.~Vera-Rodriguez, J.~Fierrez, A.~Morales, and J.~Ortega-Garcia,
  ``Deepfakes and beyond: A survey of face manipulation and fake detection,''
  \emph{Information Fusion}, vol.~64, pp. 131--148, December 2020.

\bibitem{li2020celeb}
Y.~Li, X.~Yang, P.~Sun, H.~Qi, and S.~Lyu, ``{Celeb-DF}: A large-scale
  challenging dataset for deepfake forensics,'' in \emph{Proceedings of the
  IEEE/CVF Conference on Computer Vision and Pattern Recognition (CVPR)}, 2020,
  pp. 3207--3216.

\bibitem{2022_EAAI_DeepFakes_Tolosana}
R.~Tolosana, S.~Romero-Tapiador, R.~Vera-Rodriguez, E.~Gonzalez-Sosa, and
  J.~Fierrez, ``Deepfakes detection across generations: Analysis of facial
  regions, fusion, and performance evaluation,'' \emph{Engineering Applications
  of Artificial Intelligence}, vol. 110, p. 104673, April 2022.

\bibitem{dolhansky2019deepfake}
B.~Dolhansky, R.~Howes, B.~Pflaum, N.~Baram, and C.~C. Ferrer, ``The deepfake
  detection challenge ({DFDC}) preview dataset,'' \emph{arXiv preprint
  arXiv:1910.08854}, 2019.

\bibitem{heusch2017reproducible}
G.~Heusch, A.~Anjos, and S.~Marcel, ``A reproducible study on remote heart rate
  measurement,'' \emph{arXiv preprint arXiv:1709.00962}, 2017.

\bibitem{2019_BookFFER_ContFacePAD_JHO}
J.~Hernandez-Ortega, J.~Fierrez, E.~Gonzalez-Sosa, and A.~Morales, ``Continuous
  presentation attack detection in face biometrics based on heart rate,'' in
  \emph{Video Analytics. Face and Facial Expression Recognition}, ser. LNCS,
  X.~B. et~al., Ed., vol. 11264.\hskip 1em plus 0.5em minus 0.4em\relax
  Springer, April 2019.

\bibitem{2022_Handbook_DeepFakesPhys_JHO}
J.~Hernandez-Ortega, R.~Tolosana, J.~Fierrez, and A.~Morales, ``Deepfakes
  detection based on heart rate estimation: Single- and multi-frame,'' in
  \emph{Handbook of Digital Face Manipulation and Detection}, C.~R. et~al.,
  Ed.\hskip 1em plus 0.5em minus 0.4em\relax Springer, January 2022, pp.
  255--273.

\end{thebibliography}


\end{document}